# A Reconfigurable Streaming Deep Convolutional Neural Network Accelerator for Internet of Things

Li Du, *Member, IEEE*, Yuan Du, Yilei Li and Mau-Chung Frank Chang, *Fellow, IEEE*

*Abstract*—**Convolutional neural network (CNN) offers significant accuracy in image detection. To implement image-detection using CNN in the internet of things (IoT) devices, a streaming hardware accelerator is proposed. The proposed accelerator optimizes the energy efficiency by avoiding unnecessary data movement. With unique filter decomposition technique, the accelerator can support arbitrary convolution window size. In addition, max pooling function can be computed in parallel with convolution by using separate pooling unit, thus achieving throughput improvement. A prototype accelerator was implemented in TSMC 65nm technology with a core size of 5mm². The accelerator can support major CNNs and achieve 152GOPS peak throughput and 434GOPS/W energy efficiency at 350mW, making it a promising hardware accelerator for intelligent IoT devices.**

## I. INTRODUCTION

MACHINE LEARNING offers many innovative applications in the IoT devices, such as face recognition, smart security and object detection [1-3]. State-of-the-art machine-learning computation mostly relies on the cloud servers [4-5]. Benefiting from the graph processing unit (GPU)'s powerful computation ability, the cloud can process high throughput video data coming from the devices and use CNN to achieve unprecedented accuracy on most AI applications [6]. However, this approach has its own drawbacks. Since the network connectivity is necessary for cloud-based AI applications, those applications cannot run in the areas where there is no network coverage. In addition, data transfer through network induces significant latency, which is not acceptable for real-time AI applications such as security system. Finally, most of the IoT applications have a tough power and cost budget which could tolerate neither local GPU solutions nor transmitting massive amounts of image and audio data to data center servers [7].

To address these challenges, a localized AI processing scheme is proposed. The localized AI processing scheme aims at processing the acquired data at the client side and finishes the whole AI computation without communication network access.

Conventionally, this is done through local GPU or DSP. However, this results in a limited computation ability and relatively large power consumption, making it not suitable for running computation-hungry neural network such as CNN on power limited IoT devices [8]. Consequently, it is crucial to design a dedicated CNN accelerator inside the IoT devices that can support a high performance AI computation with minimal power consumption. Some of the reported works in the neural network acceleration are focusing on providing an architecture for computing general neural network. For example, in [9], an efficient hardware architecture is proposed based on the sparsity of the neural network through pruning the network properly. However, it is a more general architecture to compute the fully-connected deep neural network without considering parameter reuse. On the contrary, the CNN has its unique feature that the filters' weights will be largely reused throughout each image during scanning. Benefiting from this feature, many dedicated CNN hardware accelerators are reported [10-12]. Most of reported CNN accelerators only focus on accelerating the convolution part while ignoring the implementation of the pooling function, which is a common layer in the CNN network. In [10], a CNN hardware accelerator using a spatial architecture with 168 processing elements is demonstrated. In [11], another dedicated convolution accelerator with loop-unfolding optimization is reported. Since pooling function is not implemented in those accelerators, the convolution results must be transferred to CPU/GPU to run pooling function and then fed back to the accelerator to compute the next layer. This data movement not only consumes much power but also limits overall performance. On the other hand, some works report highly configurable neural network processers but they require complicated data flow control. This adds hardware overhead to IoT devices. For example, [12] reports a CNN processor occupying 16 mm² silicon area in 65nm CMOS technology, which can be intolerable for low-cost IoT chips. In addition, several recent reports, such as [13], proposed to use memristors to perform neuromorphic computing for CNN. However, the fabrication of memristors currently is still not supportive in major CMOS foundry [14]. Thus this architecture is hard to embed into the IoT chips.

This paragraph of the first footnote will contain the date on which you submitted your paper for review. It will also contain support information, including sponsor and financial support acknowledgment. For example, "This work was supported in part by the U.S. Department of Commerce under Grant BS123456".

The next few paragraphs should contain the authors' current affiliations, including current address and e-mail. For example, F. A. Author is with the National Institute of Standards and Technology, Boulder, CO 80305 USA (e-mail: author@ boulder.nist.gov).

In this study, we propose a new streaming hardware architecture for CNN inference at the IoT platform and assume the CNN model is pre-trained. We focus on the optimization of the data-movement flow to minimize data access and achieve high energy efficiency for computation. A new methodology is also proposed to decompose large kernel-sized computation to many parallel small kernel-sized computations. Together with the integrated pooling function, our proposed accelerator architecture can support completed one-stop CNN acceleration with both arbitrarily sized convolution and reconfigurable pooling. The main contribution of this paper includes:

1. A CNN accelerator design using streaming data flow to achieve optimal energy efficiency.
2. An interleaving architecture to enable parallel computing for multiple output features without SRAM input bandwidth increment.
3. A methodology to decompose large-sized filter computation to be many small-sized filter computation, achieving high reconfigurability without adding additional hardware penalty.
4. A supplementary pooling block that can support pooling function while the main engine serves for CNN computation.
5. A prototype design with FPGA verification, which can achieve a peak performance of 152 GOPS and energy efficiency of 434 GOPS/W.

The paper is organized as follows. In Section II, we first introduce the main layers composing CNNs. In Section III, we introduce our system's overview architecture. In Section IV, we discuss the proposed streaming architecture to achieve high-efficiency convolution computation, filter decomposition technique to provide reconfigurability and pooling implementation. Key modules' design is explained in Section V. Finally, the experimental results are reported in Section VI and the conclusion is drawn in Section VII.

## II. Layer Description

The state-of-art CNN networks (e.g., AlexNet, VGG-18, etc.) [15-18] are mainly composed through three typical layers: convolution layer, pooling layer and classification layer. Convolution layer composes the majority of the neural network, with pooling layer inserted between two convolution layers to achieve intermediate data size reduction and non-linear mapping. Classification layer is usually included as the last layer of the CNN, which does not require a large amount of computation. Here, we assume that the classification layer can be realized through software computation and will not be implemented in the hardware accelerator.

The following subsection will explain the convolution layer and the pooling layer's functions in details.

### A. Convolution Layer

The primary role of a convolution layer is to apply convolution function to map the input (previous) layer's images to the next layer.

Since each input layer can have multiple input features (referred as channels afterward), the convolution is 3D. Unlike regular convolution, where it took the whole input data to generate one output data, the convolution in a neural network is localized through forming a regional filter window in each individual input channel. This set of the regional filter windows is regarded as one filter. The output data is obtained through computing the inner product of the filter weight and the input data covered by the filter. An output feature can be obtained by using the convolution filter to scan the input channels. Multiple output features can be computed by using different filters. In addition, a separated bias weight will be added in each final filtered result. The arithmetical representation of this function is shown as (1).

$$O[io][r][c] = B[io] + \sum_{ii=0}^{Fi-1} \sum_{i=o}^{K-1} \sum_{j=0}^{K-1} I[ii][s \times r + i][s \times c + j] \times W[io][ii][i][j]$$

$$0 \leq io < Fo, 0 \leq ii < Fi, 0 \leq r < ROW, 0 \leq c < COL \quad (1)$$

Here, $io$ represents the current output-feature's index number, $Fi$ and $Fo$ represents the total number of the input channels and output features. $r$ and $c$ represents the current output-feature's data's row and column number; $s$ is the stride size of the convolution window, $W$ represents the filter weight and $B$ represents the bias weight of each filter. $K$, $ROW$ and $COL$, are the kernel size, output-feature row size and column size respectively.

With the above parameters' definition, the input layer has $Fi$ channels. Each channel's width is $COL \times s$ and height is $ROW \times s$. The layer output includes $Fo$ features. Each feature's width is $COL$ and height is $ROW$. Filter number is same as the output-feature number. In each filter, it is constructed through $Fi$ separated filter window. Each window's kernel size is $K$. The overall convolution procedure is represented as Fig. 1.

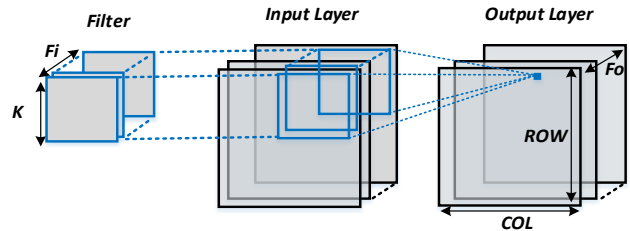

Fig.1 Example of computation of a CNN layer.

### B. Pooling Layer

In additional to the convolution layer, pooling layer is also an important part of the regular CNN. The role of the pooling layer is to extract information from a set of neighboring image pixels in each channel. Typically, the pooling layer can be separated into two categories: max pooling layer and average pooling layer. The max pooling layer selects the maximum image data's value within the pooling window, while the average pooling layer provides the average value of the data within the pooling window. The mathematical representations of these two pooling operations are defined as (2) and (3). Each input channel is pooled separately, resulting the layer's input-channel number to be equal to the output-feature number. Fig. 2 is an example of the max pooling function.

$$O\_avg[r][c] =$$

$$avg \begin{bmatrix} I[r][c] & \cdots & I[r][c+K-1] \\ \vdots & \ddots & \vdots \\ I[r+K-1][c] & \cdots & I[r+k-1][c+K-1] \end{bmatrix} \quad (2)$$

$$O\_max[r][c] =$$

$$max \begin{bmatrix} I[r][c] & \cdots & I[r][c+K-1] \\ \vdots & \ddots & \vdots \\ I[r+K-1][c] & \cdots & I[r+k-1][c+K-1] \end{bmatrix} \quad (3)$$

Here $I[r][c]$ represents the input channel's data at the position (r,c) and the kernel size of the pooling window is K.

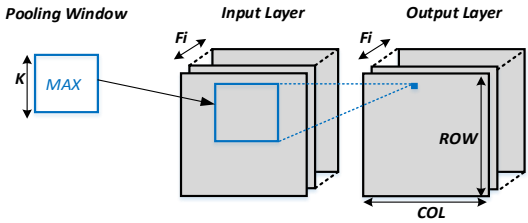

Fig.2 Example of computation of a max pooling layer.

## III. SYSTEM OVERVIEW

The overall streaming architecture of the CNN accelerator is shown in Fig. 3. It is already proved that deep networks can be represented with 16-bits fixed-point number with stochastic rounding and incur little to no degradation in the classification accuracy [19]. In addition, an implementation of the 16-bits floating adder costs much more logic gates compared to that of the 16-bits fixed-point adder [20]. Thus, the data format of this accelerator is set as the 16-bits fixed point. The accelerator includes a 96 Kbyte single port SRAM as the buffer bank to store the intermediate data and exchange data with the DRAM. The buffer bank is separated into two sets. One for the input data of the current layer and the other one is to store the output data. The input channels and output features are numbered. In each set, the buffer bank is further divided into Bank A and Bank B. Bank A is used to store the channels/features that their numbers are odd, while Bank B is used to store the channels/features that their numbers are even. In addition, a COL BUFFER module is implemented to remap the buffer bank's output to the convolution unit (CU) engine's input. The CU engine is composed of sixteen convolution units to enable highly parallel convolution computation. Each unit can support the convolution with a kernel size up to three. A pre-fetch controller is included inside the engine to periodically fetch the parameters from Direct Memory Access (DMA) controller and update the weights and bias values in the engine. Finally, an accumulation (ACCU) buffer with scratchpad is implemented in the accelerator. The scratchpad is used together with the accumulator to accumulate and store the partial convolution results coming from the CU engine. A separated max pooling module is also embedded in the ACCU buffer to pool the output-layer data if necessary.

The control of this accelerator is through 16-bit Advanced Extensible Interface (AXI) bus, the command decoder is integrated inside the accelerator. The commands for the processed CNN net are pre-stored in the DRAM in advance, and will be automatically loaded to a 128-depth command FIFO when the accelerator is enabled.

The commands can be divided into two categories: configuration commands and execution commands. Configuration commands are inserted between multiple layers to configure the upcoming layer's property, such as channel size and numbers, enable ReLU function or max pooling function. The execution commands are to initiate the convolution/pooling computation. The configuration of the shifting address value for large-sized convolution filter is also included in the execution commands (explained in Section V).

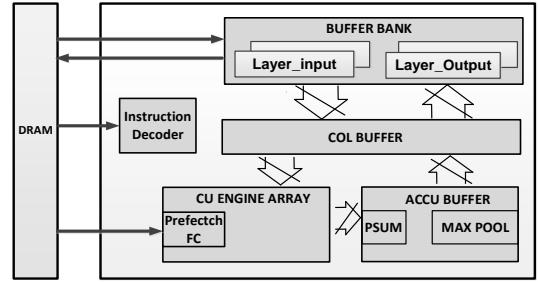

Fig.3 Overall architecture of the CNN accelerator.

The convolution begins with resetting the image scratchpad in the ACCU buffer. Then the input-layer data will be sent to CU engine sequentially. The CU engine will calculate the inner product of each channel's data with its corresponding output feature's filter's weight. Output results from the CU engine will be passed to the ACCU Buffer block and accumulated with the stored results in the scratchpad. After all the channels are scanned, the accumulated image in the scratchpad will be sent back to the Buffer bank as one of the output features.

After finishing the computation of the 1st feature, the CNN accelerator will duplicate the convolution procedure described above with updated filter weights from the DRAM, to generate the next output feature. This procedure will be continuously reproduced till all the features are calculated. The overall diagram showing this procedure is drawn as Fig. 4.

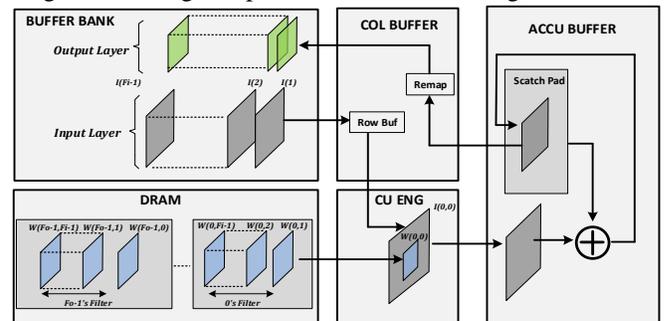

Fig.4 Convolution Computation Procedure. Input-layer data is stored by channels in the buffer bank and will be fed to the CU engine sequentially. The weight is stored in DRAM and will be fed to the CU engine during the convolution.

## IV. Streaming and Reconfigurable Features

The proposed CNN accelerator achieves reconfigurability and high energy efficiency through using three techniques below:

1. Using filter decomposition technique to support large kernel-sized filter's computation through using only 3x3-sized computation unit.

2. Streaming data flow to minimize bus control and module interface, thus reducing hardware cost while achieving high energy efficiency.

3. Separate pooling blocks to compute max pooling in parallel with convolution and reuse the convolution engine for average pooling functions to achieve minimum hardware design cost.

### A. Filter Decomposition

The filter's kernel size in a typical CNN network can range from very small size (1x1) to very large size (11x11). Hardware convolution engine is usually designed for a certain kernel size and can only support filter computation below its limited size. So when computing the convolution with kernel size above its limitation, the accelerator needs to either leave the software to do the computation or add additional hardware unit for large kernel-sized filter convolution.

To minimize the hardware resource usage, a filter decomposition algorithm is proposed to compute any large kernel-sized (>3x3) convolution through using only 3x3-sized CU. The algorithm begins with examining the kernel size of the filter. If the original filter's kernel size is not an exact multiple of three, zero padding weights will be added in the original filter's kernel boundary to extend the original filter's kernel size to be a multiple of three. Because the added weights in the boundary are 0, so the extended filter will result in same output value compared with the original filter during the computation. Next, the extended filters will be decomposed into several 3x3-sized filters. Each filter will be assigned a shift address based on its top left weight's relative position in the original filter. For example, Fig. 4 is an example of decomposing a 5x5 filter into four 3x3 filters. One row and column zero padding are added in the original filter. The decomposed filters: F0, F1, F2, F3's shift address are (0,0), (0,3), (3,0), (3,3).

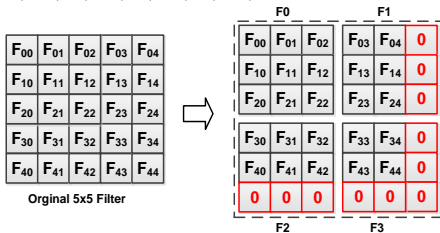

Fig.5 An 5x5 Filter decomposed into four 3x3 sub filter. F0, F1, F2, F3's shift address are (0,0), (0,3), (3,0), (3,3).

After that, we compute each decomposed filter with the input layer separately, generating several decomposed output features. Finally, we recombine those decomposed features into one final output feature through (4).

$$I_o(X, Y) = \sum_i I_{d\_i}(X + x_i, Y + y_i) \quad (4)$$

Here, $I_o$ represents the output image, $I_{d\_i}$ represents $i$'s decomposed filter's output image, $(X, Y)$ represents the current output data's coordinate address, $(x_i, y_i)$ represents $i$'s filter's shift address.

The arithmetical derivation of this filter decomposition can be described as (5)

$$F_{3K}(a, b) = \sum_{i=0}^{3K-1} \sum_{j=0}^{3K-1} f(i, j) \times I_i(a + i, b + j)$$
$$= \sum_{i=0}^{K-1} \sum_{j=0}^{K-1} \sum_{i=0}^{2} \sum_{m=0}^{2} f(3i + l, 3j + m) \times I_i(a + 3i + l, b + 3j + m)$$
$$= \sum_{i=0}^{K-1} \sum_{j=0}^{K-1} F_{3\_i\_j}(a + 3i, b + 3j) \quad (5)$$

$$F_{3\_i\_j}(a, b) = \sum_{m=0}^{2} \sum_{l=0}^{2} f(3i + l, 3j + m) \times I_i(a + 3i + l, b + 3j + m)$$
$$0 \le i < K - 1; 0 \le k < K - 1; \quad (6)$$

Here $F_{3K}(a, b)$ represents a filter with kernel size of 3K and its top-left weight is multiplied with the pixel's value at the position (a,b) in the image. Each weight in the filter is represented as $f(i, j)$ where the $(i, j)$ represents the weight's position relative to the top-left weight inside the filter and $I_i(a + 3i + l, b + 3j + m)$ represent the image pixel's value at the position of $(a + 3i + l, b + 3j + m)$ in the image. $F_{3\_i\_j}$ represents $K^2$'s different 3x3 kernel-sized filter with its computation function defined as (6). In addition, the $3i$ and $3j$ can represent as the shifting address of each 3x3 filter.

Based on (5) and (6), we can approve that a 3Kx3K filter's computation can be decomposed into $K^2$ different 3x3 filters' calculation without any loss of the computation accuracy. Fig. 6 is an example of using this filter decomposed technique to compute a 5x5 convolution.

This decomposition technique provides a benefit of maximized hardware resource usage at the penalty of adding additional zero padding in the filter boundary. Although this added zero padding results in a waste of the computation resource, the overall efficiency loss is relatively small in the CNN net. On the contrary, the CU engine design becomes much simpler as it only needs to support convolution filter size of 1x1 and 3x3. The overall efficiency loss can be computed based on the (7).

$$EL = \frac{MAC_{zero\_padding}}{MAC_{total}} \quad (7)$$

Here the $MAC_{total}$ represented the total multiply–accumulate operation the engine takes to compute a CNN network, while $MAC_{zero\_padding}$ represented the MAC operation used to compute the zero-padding part. For example, a 11x11 filter actually has $\frac{23}{144}$ MAC operation used on computing zero-padding part, resulting in an efficiency loss of 16%.

Table I is a comparison of different major CNN nets efficiency loss by using this decomposition technique.

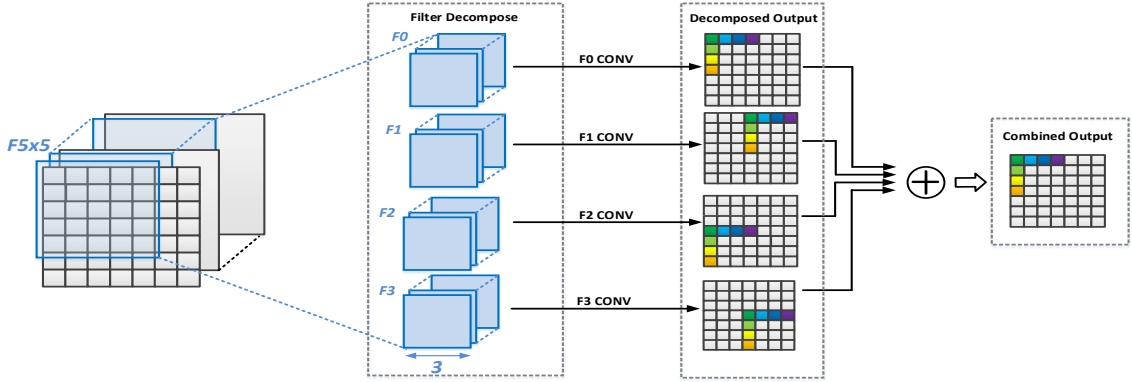

Fig. 6 Filter decomposition technique to compute a 5x5 filter on the 7x7 image. The Filter is decomposed into F0, F1, F2, F3, generating four sub-images. The sub-images are summed based on their filter's shift address. Same color's pixels will be added together to generate the corresponding pixels in the output image.

TABLE I
CONVOLUTION EFFICIENCY LOSS THROUGH DECOMPOSITION TECHNIQUE

| Net Type | Filter Kernel Size | Efficiency Loss |
|---|---|---|
| AlexNet | 3-11 | 13.74% |
| ResNet-18 | 1-7 | 1.64% |
| ResNet-50 | 1-7 | 0.12% |
| Inception V3 | 1-5 | 0.89% |

As TABLE I shows, AlexNet exhibits the largest Efficiency Loss since it has a large 11x11 filter in the first layer. On the contrary, small filter-sized nets such as Resnet-18, Resnet-50, Inception V3, have a very small efficiency loss due to zero-padding. Hence, they are well suited for this architecture.

### B. Streaming Architecture

To minimize the data movement and achieve optimal energy efficiency for the convolution computation, a streaming architecture is proposed for the CNN accelerator. For a regular CNN convolution, it includes multiple levels of data and weights reuse:

1. Every set of the filter weights is reused to scan the whole channel's image.
2. Every output feature is generated through scanning the same input layer.

The streaming architectures reduce the data movement through benefiting the above-listed features in CNN convolution.

#### 1) Filter Weight Reuse:

In each filter, the weights between kernels are different. Each kernel's weights will only be used with the particular input channel's data. To benefit from this, all the filter weights are stored in the DRAM and will only be fetched into the accelerator during the convolution.

During the 3x3 convolution, the fetched filter weights will be stored in the CU engine and input channel's image data will stream into the CU engine. The CU engine will produce the inner product between the weights and the streamed-in data, generating a corresponding output feature's partial result to the ACCU buffer for accumulation. The weights in the CU engine will not be updated until the whole channel is scanned. The 1x1 convolution follows the similar approach as the 3x3 convolution except that seven out of nine multipliers are turned off during the convolution. The left two multipliers will be turned on to calculate two different output features' partial summation result simultaneously.

Fig. 7 is an example shown one filter window movement of this flow. The real implementation includes sixteen 3x3 filter windows to process multiple rows' data simultaneously. By using this filter window to scan the input channel, the data flow and module interface become much simpler and hence the hardware design cost is reduced.

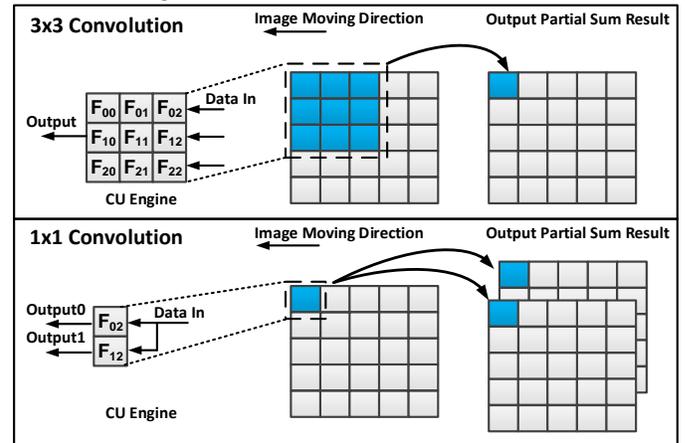

Fig. 7 Data flow of the streaming architecture.

The output bandwidth of the Buffer Bank is set to be 256 bits/cycle with each data size as 16 bits, corresponding to stream in sixteen data from different rows to the CU engine simultaneously. The sixteen data are divided into two sets: eight data are from the channels that their numbers are odd and the other eight data is from the channels that their numbers are even.

To maximize the usage of the buffer bank's output bandwidth, a two-rows' FIFO buffer is paired with each set of the row data, transferring the eight input rows to ten overlapping output rows. This enables running eight 3x3 CU in parallel for each set of the row data. The FIFO buffer included in the COL buffer is shown in Fig. 8. Here we only draw half sized COL buffer for the even-number channels' data. Real implementation includes the FIFO buffer for both even and odd channels.

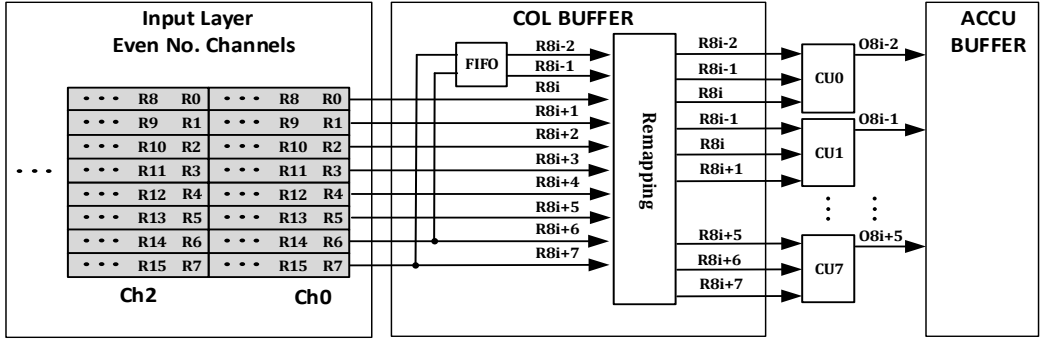

Fig.8 Half of COL Buffer Architecture of the CNN accelerator, input channels are stored sequentially in the buffer. Each channel is stored by rows, $Ri$ represents $i$'s row's data.

### 2) Input Channel Reuse:

In the 1x1 convolution, each output-feature data computation only requires one multiplication in each channel. This results in wasting a majority of hardware resource as most of the multipliers in the CU engine will not be used.

To accelerate the computation in the 1x1 convolution, an interleaving architecture is proposed to compute two output features' results in parallel in the 1x1 convolution. Since computation of each output feature requires scanning the same input layer, the accelerator can compute multiple output features simultaneously during one scanning. However, if multiple features are computed simultaneously, it will result in a proportional output bandwidth increment for the CU engine. For example, output two features simultaneously will lead the CU engine to generate twice output data bandwidth compared to its input data bandwidth.

To prevent this, an interleaving architecture is proposed through separating the sixteen input data into even-number channel's data and odd-number channel's data. These two sets of data are individually multiplied with two different features' weights, resulting in a total of 32 data (two output features' partial results) at the CU engine output. However, since the 32 data streams are generated from two different channels, a summation function is required at the CU output to combine the same feature's partial results from different channels. By doing this, the data bandwidth is reduced by half and hence the final output of the adder will be same as the input data bandwidth. The details implementation of this function is drawn in Fig. 9. Here the $X(0,0) \cdots X(0,7)$ represents the $1^{st}$ to the $8^{th}$ row's data of the odd-number channels and the $X(E,0) \cdots X(E,7)$ represents the $1^{st}$ to the $8^{th}$ row's data of the even-number channels. $O(0,1), E(0,1)$ is the $1^{st}$ features' partial result from odd-number and even-number channels and $O(0,2), E(0,2)$ is the $2^{nd}$ features' partial result from the odd-number and even-number channels.

### 1x1 Convolution

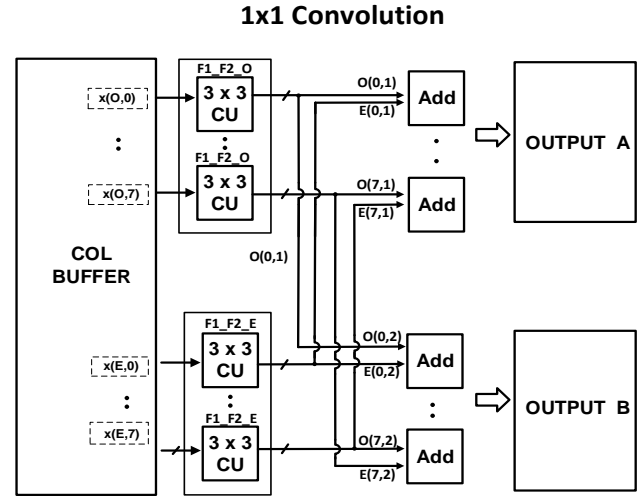

Fig. 9 Data flow of the streaming architecture in the 1x1 convolution mode.

### C. Pooling

Pooling functions are also implemented in the accelerator. The pooling functions can be separated into two categories: max pooling and average pooling.

### 1) Average Pooling:

To minimize hardware cost, the average pooling function is implemented through reusing the convolution engine. This can be achieved through replacing the average pooling layer with the same kernel-sized convolution layer using the following steps:

1. Create a convolution layer with the output features' number to be equal to the input channels' number. The kernel size is same as the pooling window size
2. In each filter, set the corresponding channel's filter's weight to $\frac{1}{K^2}$, where K is the kernel size. All other channels' filter weights are set to 0.

The arithmetical representation of this convolution layer can be derived as (8).

$$O[io][r][c] =$$

$$\sum_{ii=0}^{I} \sum_{i=0}^{K-1} \sum_{j=0}^{K-1} I[io][ii][r+i][c+j] \times W[io][ii][i][j]$$

$$W[io][ii][i][j] = \begin{cases} \frac{1}{K^2} & if\ ii = io \\ 0 & if\ ii \neq io \end{cases} \quad (8)$$

Here $ii$ and $io$ are the input-channel number and output-feature number, $r$ and $c$ are the output feature's row and column's number, $W$ represents the weight of the filter. $K$ is the kernel size of the average pooling window.

### 2) Max Pooling:

The max pooling layer is implemented as a separate block inside the ACCU buffer and it is used to pool the output feature coming from the convolution block. The pooling block is designed to support pooling window size of two and three, which covers major CNNs [15-18]. The detailed implementation of this block and its connection to the scratchpad will be described in Section V.

## V. MODULE IMPLEMENTATION

In this sections, three major modules: CU engine, ACCU buffer and Max pooling in this accelerator will be discussed.

### A. CU Engine

As described in Section IV, the accelerator uses nine multipliers to form a CU and sixteen CUs to compose a CU engine. The module implementation of the CU is shown in Fig. 10.

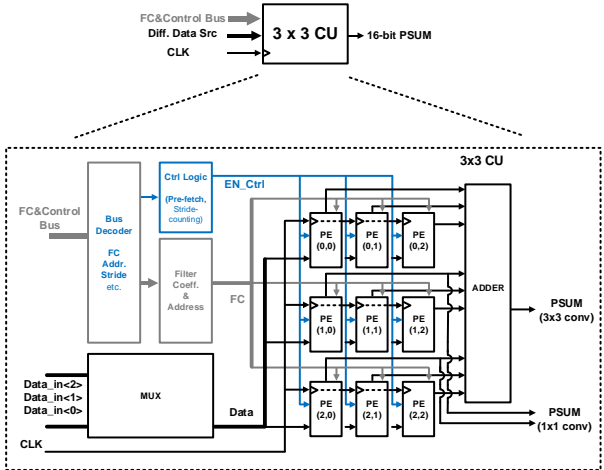

Fig. 10 Implementation of the 3x3 CU engine.

The CU engine array includes nine processing engines (PE) and an adder to combine the output. The PE provides a multiplication function for its input data and the filter's weight and meanwhile passes its input data to the next stage's PE through a D flip-flop. The multiplication function can be turned on/off based on the EN_Ctrl signal to save the computation power when convolution stride size is larger than one.

In the 3x3 convolution, the multiplied result will send to the adder in the CU to perform the summation and deliver the summed result to the final output. Filter weights will be fetched from the DRAM through the DMA controller and pre-stored in the CU through a global bus. When one channel is scanned, a synchronized filter updated request signal will be sent to the CU to update the filter weights at the PE's input for the upcoming channel.

In the 1x1convolution, only PE (1,0) and PE (2,0) will be turned on. The adder will be disabled and the two output results are directly fed out as the two output-feature partial results.

### B. ACCU BUFFER

The ACCU Buffer is used to accumulate the output partial summation results from the CU engine and meanwhile temporary store the feature output data in its scratchpad, waiting for the buffer bank to read back. The ACCU buffer includes a ping-pong buffer as the scratchpad, an accumulator to sum the partial result, a separate pooling block for max pooling and a readout block to read data from the scratchpad back to the buffer bank.

The ping-pong buffer is separated into two different sub-buffers. During the convolution, only one buffer will be pointed to the accumulator while the other buffer will be connected to the pooling blocks and the readout blocks. This enables the core to process the pooling functions and the convolution functions simultaneously. In addition, reading data from the scratchpad back to the buffer bank can also be processed in parallel with the convolution.

When the accumulator finished accumulating one output feature, the ping-pong buffer will switch its sub-buffers directions, pointing the buffer that stores the output feature to the pooling blocks and the readout blocks. Meanwhile, the sub buffer which previously connected to the pooling side will turn to the accumulator to continuously accumulate the next output-feature partial summation result. In addition, the ReLU function is implemented during the readout. The ReLU function can be realized through zeroing the negative output from the readout blocks.

Compared with the convolution, the readout and pooling functions only need to scan one output feature each time, resulting a much shorter time to process. Benefited from this, the accelerator can continuously run convolution without any speed loss on the pooling and the data readout. The detailed implementation of the ACCU Buffer architecture is shown as Fig.11.

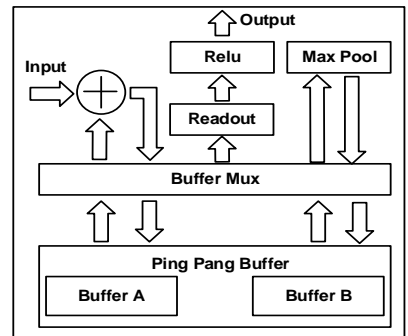

Fig. 11 The ACCU Buffer includes a Ping-Pang buffer formed by the Buffer A and the Buffer B. The two buffers will be switched back and forth between the accumulator and the Readout/Max pool blocks to enable parallel processing.

## C. Max Pool

Fig. 12 shows an overview architecture of the max pooling module and its connection to the scratchpad. The scratchpad stored eight rows' data from one output feature in parallel. The eight rows' data share one column address and can be accessed simultaneously. Because of the stride size's difference in the convolution, data stored in the scratchpad may not be all validated. For example, when the stride is equal to 2, only R0, R2, R4, R6 store the validate data. In addition, the pool window's kernel size can also be configured to be 2 or 3.

To accommodate different convolution strides and pool-size cases, a MUX is put in front of the max pooling module to select the validated input data to the corresponding max-pool units. The max-pool unit is implemented with a four-input comparator and a feedback register to store the intermediate comparator output result. In addition, an internal buffer is embedded in the max pooling module. This is to buffer the intermediate results if some of the data inside the pooling window are not ready.

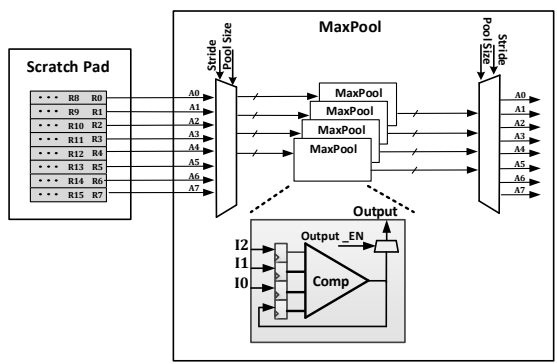

Fig. 12 Overall architecture of the Maxpooling module, Ri represents row i's data. The pooled output will be fed back to the scratchpad.

When a pooling begins, the comparator first takes three input data coming from nearby rows (two data in 2x2 case) and output the maximum value among the input data. This temporary maximum value will be fed back to the comparator's input and regarded as one additional input to compare with the next clock cycle's input data. This procedure will be duplicated till the whole pooling window's input data is scanned. After that the output enabling signal will be validated and output the maximum value in the pooling window.

## VI. RESULTS

The accelerator was implemented in TSMC 65nm technology and the layout characteristics of the accelerator are shown in Fig. 13. The core dimension is 2mm x 2.5mm and achieves a peak throughput of 152 GOP/s at a 500MHz core clock. Since the core can support both arbitrary sized convolution layer and the pooling function, it can be used to accelerate major CNNs. A summary of the chip specifications is listed in Table II. The power is based on the synthesis report from the Synopsis Design Compile, while the area and clock speed are based on Place&Route report in Cadence. Here, PE is representing the processing engine in the chip which is a multiplier in each CU. The energy-efficiency is defined as the peak throughput divided by the dynamic power consumption.

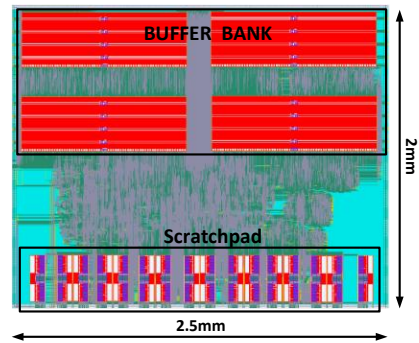

Fig. 13 Layout view of the accelerator.

### TABLE II
#### PERFORMANCE SUMMARY

| Technology | TSMC 65nm RF 1P6M |
|---|---|
| Supply Voltage | 1V |
| Clock Rate | Up to 500MHz |
| Dynamic Power Consumption | 350mW @ 500MHz |
| Core Area | 2mm x2.5mm |
| Gate Count | 1.3M |
| Number of PEs | 144 |
| On-Chip Single Port SRAM | 96 K bytes |
| Scratch Pads Memory | 16K byte |
| Peak Throughput | 152GOPS |
| Energy Efficiency | 434GOPS/W |
| Arithmetic Precision | 16-bit fixed-point |
| Supported CNN feature | filter kernel size:1-23 |
| | Num. of filters:1-1024 |
| | Num. of channels:1-1024 |
| | Num. of features:1-1024 |
| | Stride Size : 1, 2, 4 |
| Supported Pooling feature | Avg Pool Size: 1-23 |
| | Max Pool Size:2,3 |

The area breakdown of the accelerator is shown in Fig. 14. The area estimation includes the logic cells, registers, and single port/dual port SRAMs generated by the ARM compiler. As it shows, the CU engine only occupies 17% of the total area. The majority of the area is occupied by the buffer bank and scratchpad. The scratchpad is designed using dual port SRAM to support the continuous streaming while buffer bank is implemented as single port SRAM. Although the scratchpad memory size is only 1/6 compared to the buffer bank, it is still occupied more than half of the buffer bank's area.

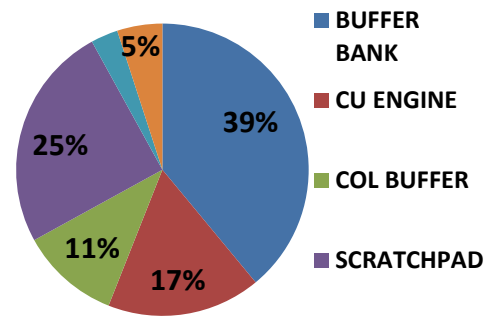

Fig. 14 Area breakdown of the accelerator.

TABLE III
TRAFFIC SIGN CNN ARCHITECTURE

| Layer | Type | Channel Size | Channel No. | Kernel Size | Stride |
|-------|------|--------------|-------------|-------------|--------|
| 1 | Input | 32 x 32 | 3 | — | — |
| 2 | convolution | 32 x 32 | 3 | 5x5 | 1 |
| 3 | max pooling | 32 x 32 | 64 | 2x2 | 2 |
| 4 | convolution | 16 x 16 | 64 | 5x5 | 1 |
| 5 | max pooling | 16 x 16 | 16 | 2x2 | 2 |
| 6 | convolution | 8 x 8 | 16 | 5x5 | 1 |
| 7 | fully-connected | 8 x 8 | 16 | — | — |

To verify the performance of the accelerator, we have downloaded the hardware accelerator IP into the Xilinx Zynq-7200 FPGA and demonstrate the core's functions using modified LeNet-5 [21] to detect the traffic sign. The filter's weights are fetched from the DRAM through the FPGA's existing DMA controller. The DMA controller is configured as 256-depth 64-bits width. The traffic-sign net includes three convolution layers and two pooling layer and its architecture is summarized in Table III.

The application processor (AP) integrated into the FPGA is used to control the accelerator and initiate the computation. Through using the DMA controller inside the FPGA, the accelerator can successfully access the data and the weights stored in the DRAM. The demonstration setup is shown in Fig.15. The demonstration begins with downloading a traffic sign into the FPGA. After the computation, the detected traffic sign result will be sent back to the PC and display on the monitor. A raw video demonstration is shown in [22].

Even the demonstrated LeNet-5 Model only has an input channel size of 32×32, this accelerator can fit for channel size that larger than this. In fact, a large-sized channel can improve the energy-efficiency of the system. This is due to the fact that large-sized channels lead to more filter-weights reuse during the computation. For example, a 100×100 input channel will result in approximately 10000 times filter reuse during the scanning of the image, while a 10×10 input channel only has 100 times filter-weights reuse.

When the input channel or intermediate data size is larger than the total available SRAM size. A DMA controller is needed to exchange data between DRAM and on-chip SRAM. This will cost a large energy consumption as the intermediate data is exchanged between the DRAM and the on-chip SRAM.

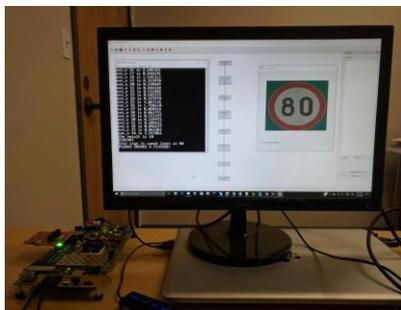

Fig. 15 Traffic Sign Demonstration on the Xilinx Zynq-7200 FPGA.

In addition, the data format in this hardware accelerator is set to be 16-bits fixed point to achieve minimized hardware cost. Through re-designing the multiplier and adder in the CU block, this architecture can also be used with other data formats such as 16-bits floating point, 32-bits floating points or 8-bits fixed point.

Table IV is a comparison of the designed accelerator with other reported work. As it shows, this accelerator achieves high energy efficiency and comparable performance with low area cost, making it suitable to be integrated into the IoT devices.

TABLE IV
PERFORMANCE COMPARISON

| | This work | [9] | [10] |
|---|-----------|-----|------|
| Core Area | 5mm² | 12mm² | 16mm² |
| Peak Throughput | 154GOPS | 84GOPS | 64GOPS |
| Gate Count | 1.3M | 1.2M | 3.2M |
| Supply Voltage | 1V | 0.82-1.17V | 1.2V |
| Peak Throughput | 152GOPS | 84GOPS | 64GOPS |
| Energy Efficiency | 434GOPS/W | 166GOPS/W | 1.4TOPS/W |
| Technology | 65nm | 65nm | 65nm |
| Precision | 16-bit | 16-bit | 16-bit |
| MaxPool Support | Yes | No | Yes |
| AvgPool Support | Yes | No | No |

## VII. CONCLUSION

In this paper, we propose a streaming architecture for the CNN hardware accelerator. The proposed accelerator optimizes the energy efficiency by reducing unnecessary data movement. It also supports arbitrary window sized convolution by using filter decomposition technique. In addition, pooling function is also supported in this accelerator through integrating separate pooling module and proper configuration of the convolution engine. The accelerator is implemented in TSMC 65nm technology with a core size of 5mm². A traffic-sign net is implemented using this hardware IP and verified on the FPGA. The result shows that this accelerator can support most popular CNNs and achieve 434GOPS/W energy efficiency, making it suitable to be integrated with the IoT devices.